\documentclass[sigconf, screen]{acmart}
\setcopyright{none}
\AtBeginDocument{%
  \providecommand\BibTeX{{%
    \normalfont B\kern-0.5em{\scshape i\kern-0.25em b}\kern-0.8em\TeX}}}

\copyrightyear{2025}
\acmYear{2025}
\setcopyright{acmlicensed}\acmConference[MM '25]{Proceedings of the 33rd ACM International Conference on Multimedia}{October 27--31, 2025}{Dublin, Ireland}
\acmBooktitle{Proceedings of the 33rd ACM International Conference on Multimedia (MM '25), October 27--31, 2025, Dublin, Ireland}
\acmDOI{10.1145/3746027.3755027}
\acmISBN{979-8-4007-2035-2/2025/10}

\usepackage{algorithm}
\usepackage{algorithmic}
\usepackage{mathrsfs}
\usepackage{marvosym}
\usepackage{multirow}
\usepackage{graphicx}
\usepackage{subfigure}
\usepackage{adjustbox}
\usepackage{colortbl}  

\begin{document}

\title{CopyJudge: Automated Copyright Infringement Identification and Mitigation in Text-to-Image Diffusion Models}

\author{Shunchang Liu}
\authornotemark[1]
\affiliation{%
  \institution{EPFL}
  \city{Lausanne}
  \country{Switzerland}}
\email{shunchang.liu@epfl.ch}

\author{Zhuan Shi}
\authornote{indicates equal contribution.}
\authornote{indicates corresponding author.}
\affiliation{%
  \institution{Mila - Quebec AI Institute \\
   Mcgill University}
  \city{Montreal}
  \country{Canada}}
\email{zhuan.shi@mila.quebec}

\author{Lingjuan Lyu}
\affiliation{%
  \institution{Sony AI}
  \city{Zurich}
  \country{Switzerland}}
\email{lingjuan.lv@sony.com}

\author{Yaochu Jin}
\affiliation{%
  \institution{Westlake University}
  \city{Hangzhou}
  \country{China}}
\email{jinyaochu@westlake.edu.cn}

\author{Boi Faltings}
\affiliation{%
  \institution{EPFL}
  \city{Lausanne}
  \country{Switzerland}}
\email{boi.faltings@epfl.ch}

\begin{abstract}
Assessing whether AI-generated images are substantially similar to copyrighted works is a crucial step in resolving copyright disputes. In this paper, we propose \textit{CopyJudge}, an automated copyright infringement identification framework that leverages large vision-language models (LVLMs) to simulate practical court processes for determining substantial similarity between copyrighted images and those generated by text-to-image diffusion models. Specifically, we employ an abstraction-filtration-comparison test framework with multi-LVLM debate to assess the likelihood of infringement and provide detailed judgment rationales.
Based on the judgments, we further introduce a general LVLM-based mitigation strategy that automatically optimizes infringing prompts by avoiding sensitive expressions while preserving the non-infringing content. Besides, our approach can be enhanced by exploring non-infringing noise vectors within the diffusion latent space via reinforcement learning, even without modifying the original prompts.
Experimental results show that our identification method achieves comparable state-of-the-art performance, while offering superior generalization and interpretability across various forms of infringement, and that our mitigation method could more effectively mitigate memorization and IP infringement without losing non-infringing expressions\footnote{Code is available at \url{https://github.com/shunchang-liu/CopyJudge}.}.

\end{abstract}


\ccsdesc[500]{Security and privacy~Social aspects of security and privacy}
\ccsdesc[500]{Computing methodologies~Machine learning}

\keywords{Copyright Protection, Text-to-image Diffusion Model}
\maketitle

\section{Introduction}
\label{introduction}
Text-to-image generative models \cite{Rombach_2022_CVPR, Betker2023ImprovingImageGeneration, team2023gemini, esser2024scaling, hurst2024gpt, zhang2023adding, zhang2023text, hintersdorf2024finding} have transformed creative industries by producing detailed visuals from text prompts. However, these models have been found to sometimes memorize and reproduce content from their training data \cite{carlini2023extracting,somepalli2023diffusion, ren2024copyright, wang2024replication, shi2024rlcp, shi2024copyright, zhang2024forget}. This raises significant concerns about copyright infringement, especially when the generated images closely resemble existing copyrighted works. According to U.S. law \cite{roth_greeting_cards_wikipedia}, also referenced by most countries, a work can be considered infringing if it constitutes \textit{substantial similarity} to another work. Therefore, determining whether AI-generated images infringe on copyright requires a clear and reliable method to compare them with copyrighted materials to identify substantial similarity.

However, identifying substantial similarity is not a trivial task. There are already some methods to assess image similarity through distance-based metrics, e.g., $L_2$ norm \cite{carlini2023extracting}. However, we found that these manually designed metrics do not always align with the human judgment for infringement determination. Additionally, it often suffer from insufficient generalization ability and lack interpretable results. This motivates the need for an approach that better measures substantial similarity, one that is more \textit{human-centered}, \textit{interpretable}, and \textit{generalized} to handle copyright infringement identification in AI-generated images.


Recently, large-scale models have already been successfully applied as judges in fields such as finance, education, and healthcare \cite{gu2024survey, li2024llms, zhuge2024agent}. In this paper, we attempt to leverage large vision-language models (LVLMs) to model the practical court decisions on substantial similarity. However, directly applying large models for infringement identification may face unreliable outputs due to their limited comprehension or potential misinterpretation \cite{xu2025can}. To address this, we propose \textit{CopyJudge}, an automated abstraction-filtration-comparison framework with multi-LVLM debate to reliably follow the court decision process on identifying substantial similarity. 

Specifically, referring to software abstraction test \cite{abramson2002promoting}, we decompose the image into different layers or elements, such as composition, color, and theme, to distinguish between the basic concepts of the image and its specific expressions. Then, we filter out parts that are not copyright-protected, such as public and functional expressions. Finally, we compare the filtered portions to assess whether there is substantial similarity. To enhance the reliability of the judgment, we employ a \textit{multi-agent debate} \cite{du2023improving, chan2023chateval} method where multiple LVLMs discuss and score the similarity. Each LVLM can make judgments based on the scores and reasons provided by other LVLMs. Ultimately, another LVLM-based meta-judge gives the final score and rationale based on the consensus of the debate. To enhance the consistency with human preferences, we inject human priors into each LVLM via few-shot demonstrations \cite{agarwal2024many}. 


Given the judging results, we further explore how to mitigate infringement issues in text-to-image diffusion models. Utilizing our \textit{CopyJudge}, we propose an easy but effective black-box infringement mitigation strategy that leverages a defense LVLM to iteratively optimize the input infringing prompts. This process avoids generating sensitive infringing expressions by querying supporting infringement rationales, while preserving the integrity of the original content. Moreover, if the input latent of the diffusion model is controllable, we could further enhance the mitigation approach by exploring specific non-infringing noise vectors within the latent space in a reinforcement manner, with the reward being reducing the predicted infringement score. This helps avoid infringement while maintaining the desired output characteristics, even without changing the original prompts. 


In summary, our contributions are as follows:
\begin{itemize}
\item We propose \textit{CopyJudge}, an automated abstraction-filtration-comparison framework powered by a multi-LVLM debate mechanism, designed to efficiently detect copyright-infringing images generated by text-to-image diffusion models. 
\item Given the judgment, we introduce an adaptive mitigation strategy that automatically optimizes prompts and explores non-infringing latent noise vectors of diffusion models, effectively mitigating copyright violations while preserving non-infringing expressions.
\item Extensive experiments demonstrate that our identification method matches state-of-the-art performance, with improved generalization and interpretability, while our mitigation approach more effectively prevents infringement without losing non-infringing expressions.
\end{itemize}

\section{Related Work}
\label{related}
\textbf{Copyright infringement in text-to-image models.}
Recent research \cite{carlini2023extracting, somepalli2023diffusion, somepalli2023understanding, gu2023memorization, wang2024replication, wen2024detecting, chiba2024probabilistic} highlights the potential for copyright infringement in text-to-image models. These models are trained on vast datasets that often include copyrighted material, which could inadvertently be memorized by the model during training \cite{vyas2023provable, ren2024copyright}. Additionally, several studies have pointed out that synthetic images generated by these models might violate IP rights due to the inclusion of elements or styles that resemble copyrighted works \cite{poland2023generative, wang2024stronger}. Specifically, models like stable diffusion \cite{Rombach_2022_CVPR} may generate images that bear close resemblances to copyrighted artworks, thus raising concerns about IP infringement \cite{shi2024rlcp}. 

\textbf{Image infringement detection and mitigation.}
The current mainstream infringing image detection methods primarily measure the distance or invariance in pixel or embedding space \cite{carlini2023extracting, somepalli2023diffusion, shi2024rlcp, wang2021bag, wang2024image}. For example, \citeauthor{carlini2023extracting} uses the $L_2$ norm to retrieve memorized images. \citeauthor{somepalli2023diffusion} use SSCD \cite{pizzi2022self}, which learns the invariance of image transformations to identify memorized prompts by comparing the generated images with the original training ones. \citeauthor{zhang2018unreasonable} compare image similarity using the LPIPS distance, which aligns with human perception but has limitations in capturing certain nuances. \citeauthor{wang2024image} transform the replication level of each image replica pair into a probability density function. Studies \cite{wen2024detecting, wang2024evaluating} have shown that these methods have lower generalization capabilities and lack interpretability because they do not fully align with human judgment standards. For copyright infringement mitigation, the current approaches mainly involve machine unlearning to remove the model's memory of copyright information \cite{bourtoule2021machine, nguyen2022survey, kumari2023ablating, zhang2024forget} or deleting duplicated samples from the training data \cite{webster2023duplication, somepalli2023understanding}. These methods often require additional model training. On the other hand, \citeauthor{wen2024detecting} have revealed the overfitting of memorized samples to specific input texts and attempt to modify prompts to mitigate the generation of memorized data. Similarly, \citeauthor{wang2024evaluating} leverage LVLMs to detect copyright information and use this information as negative prompts to suppress the generation of infringing images.

\section{LVLM for Infringement Identification}
\label{identification}
\begin{figure*}[!ht]
    \centering
    \includegraphics[width=\linewidth]{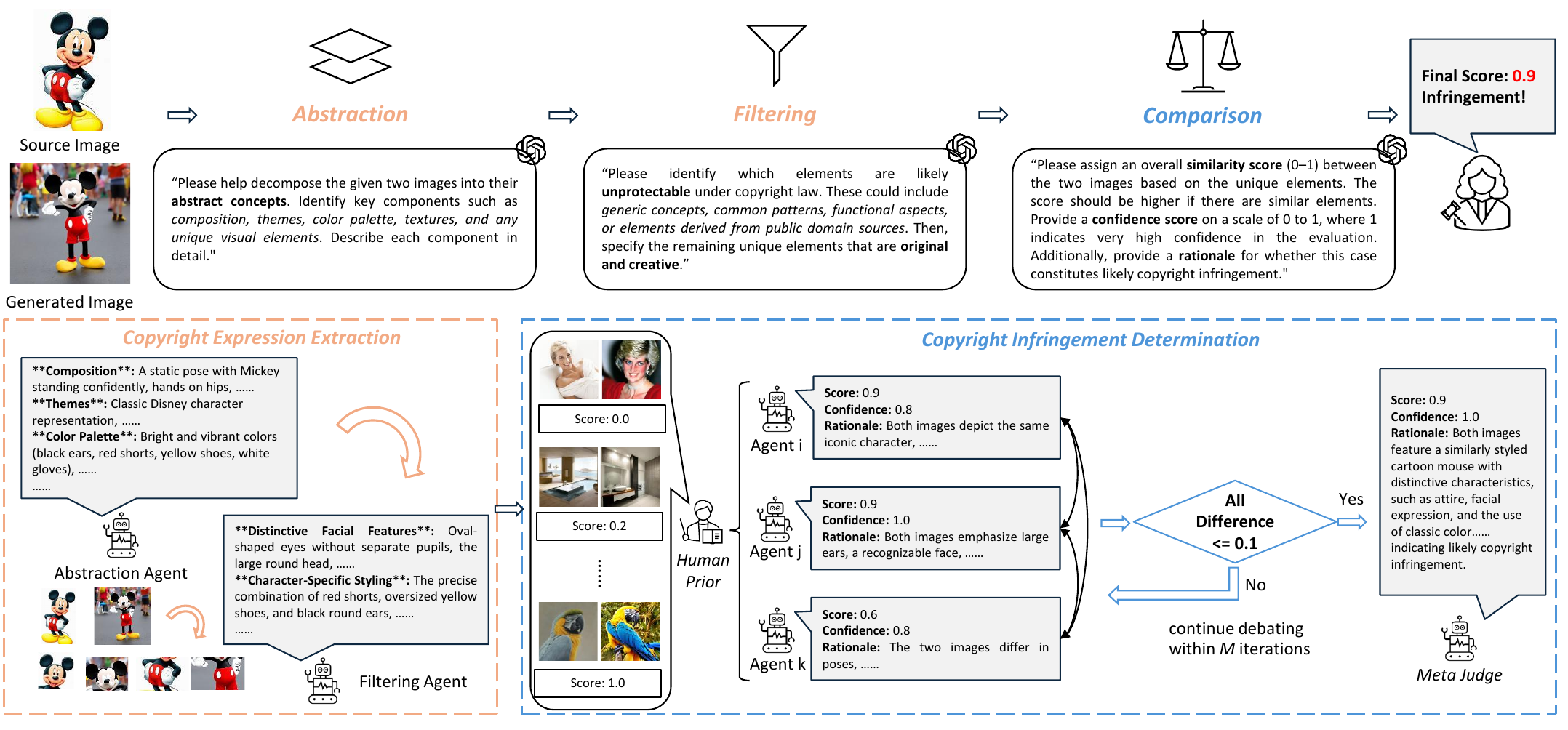}
    \vspace{-3mm}
    \caption{CopyJudge: an automated abstraction-filtering-comparison framework for image copyright infringement identification. This LVLM-based framework automatically decides whether a generated image infringes copyright by combining copyright expression extraction with infringement determination through multi-LVLM debate.}
    \label{fig:iden}
\end{figure*}

\subsection{Problem Formulation}
\textbf{Use case.} Our goal is to determine whether an image infringes the copyright of a known copyrighted image, where a copyright holder identifies a potentially infringing image and seeks to evaluate its similarity to their copyrighted work. In such scenarios, the copyright owner inherently possesses prior knowledge of the source material, making the requirement to specify it legally and practically justified. This aligns with real-world copyright enforcement workflows, where disputes often arise after a rights holder detects a suspected infringement and initiates a formal assessment. 

Based on U.S. law \cite{roth_greeting_cards_wikipedia} and similar laws in other countries, given an image $x$ created with access to the copyrighted image $x_{cr}$, if $x$ and $x_{cr}$ exhibit \textit{substantial similarity}, then $x$ is deemed to infringe the copyright of $x_{cr}$. Motivated by this, we aim to establish a substantial similarity identification model $f$, which takes 
$x$ and $x_{cr}$ as inputs and outputs a similarity score $s$. When $s$ exceeds a threshold $\gamma$, we determine that $x$ infringes on $x_{cr}$. This can be defined as:
\begin{equation}
\text{IsInfringement}(x) = \mathbb{I}(f(x, x_{{cr}}) > \gamma),
\end{equation}
where $x$ represents the AI-generated image, and $x_{cr}$ is the corresponding copyrighted image. 


\subsection{Abstraction-Filtering-Comparison Framework}
For the process of identifying substantial similarity, we refer to the \textit{abstraction-filtering-comparison} test method \cite{abramson2002promoting}, which has been widely adopted in practical court rulings on infringement cases, and propose an automated infringement identification framework using large vision-language models, as seen in Figure \ref{fig:iden}. In the \textit{copyright expression extraction stage}, we break down images into different elements (such as composition and color patterns), and filter out non-copyrightable parts, leaving copyrighted portions to assess substantial similarity. In the next \textit{copyright infringement determination} stage, multiple LVLMs debate and score the similarity of images given the copyrighted elements, with a final decision made by a meta-judge LVLM based on their consensus. Human priors are injected into the models through few-shot demonstrations to better align with human preferences. 

\textbf{Copyright expression extraction via image-to-text abstraction and filtration.} The process of distinguishing between the fundamental ideas and the specific expressions of an image is a crucial step in determining copyright protection. The core idea of our method is to break down the image into different layers or components, in order to examine the true copyright elements.

First, during the abstraction phase, the image is analyzed and decomposed into its fundamental building blocks. This involves identifying the core elements that contribute to the overall meaning or aesthetic of the image, such as composition, themes, color palette, or other unique visual elements. We can implement this using an LVLM $\pi_{abs}$, defined as:
\begin{equation}
\pi_{abs}(x, x_{{cr}}, p_{abs}) \to (z, z_{cr}),
\end{equation}
where $z$ and $z_\text{cr}$ represent the expressions of $x$ and $x_{{cr}}$ in text after decoupling, respectively.
The goal is to abstract away the superficial features of the image that do not hold significant creative value and instead focus on the underlying concepts that convey the essence of the work.

The next step is filtering. At this stage, elements of the image that are not eligible for copyright protection are removed from consideration. These can include generic concepts, common patterns, functional aspects, or elements derived from public domain sources.
For example, standard design patterns or commonly used motifs in artwork may not be deemed original enough to warrant protection under copyright law. This process could be defined as:
\begin{equation}
\pi_{fil}(z, z_{{cr}}, p_{fil}) \to (z^c, z^c_{cr}),
\end{equation}
where $z^c$ and $z^c_{cr}$ are the filtered copyright expressions, and $\pi_{fil}$ is another independent LVLM. Filtering helps ensure that only the truly creative, original aspects of the image are preserved for comparison.

The following step is to conduct a comparison of the remaining abstracted elements to assess the degree of similarity. This process helps determine whether the image in question constitutes a derivative work or if it has enough original expression to qualify for copyright protection. To ensure the reliability of the results, we used multi-LVLM debates to perform the comparison and make the final infringement determination.

\begin{algorithm}
\caption{CopyJudge: Automated Copyright Infringement Identification via Abstraction-Filtering-Comparison Framework}
\label{alg:infringement_identification}
\begin{algorithmic}[1]
\REQUIRE Image pair $(x, x_{cr})$, LVLMs $\{\pi_\text{abs}, \pi_\text{fil}, \pi_1, \dots, \pi_N, \pi_f\}$, thresholds $\alpha, \gamma$, maximum debate rounds $M$, human reference dataset $D_h$, score set $S_h$
\STATE \textbf{Stage 1: Copyright Expression Extraction}
\STATE Extract abstracted expressions: \newline
    \quad $(z, z_{cr}) \leftarrow \pi_\text{abs}(x, x_{cr}, p_\text{abs})$
\STATE Filter non-copyrightable elements: \newline
    \quad $(z^c, z^c_{cr}) \leftarrow \pi_\text{fil}(z, z_{cr}, p_\text{fil})$

\STATE \textbf{Stage 2: Copyright Infringement Determination}
\FOR{each LVLM $\pi_i$, where $i \in \{1, 2, \dots, N\}$}
    \STATE Compute similarity score, confidence, and rationale:
    \quad $(s_i, c_i, r_i) \leftarrow \pi_i(x, x_{cr}, z^c, z^c_{cr}, p_i, D_h, S_h)$
\ENDFOR

\STATE Initialize debate round counter $m \gets 1$
\WHILE{$m \leq M$}
    \FOR{each LVLM $\pi_i$}
        \STATE Receive responses $\{(s_j, c_j, r_j) | j \neq i\}$
        \STATE Update score based on peer evaluations
    \ENDFOR
    \STATE Check consensus: $\left| s_i - s_j \right| \leq \alpha, \forall i, j$
    \IF{consensus achieved}
        \STATE Break loop
    \ENDIF
    \STATE $m \gets m + 1$
\ENDWHILE

\STATE Meta-judge LVLM final decision:
\STATE $(s_f, c_f, r_f) \leftarrow \pi_f(x, x_{cr}, z^c, z^c_{cr}, p_f, D_h, S_h, S_m, C_m, R_m)$

\STATE \textbf{Determine infringement:}
\STATE $\text{IsInfringement}(x) \gets \mathbb{I}(s_f > \gamma)$

\end{algorithmic}
\end{algorithm}

\textbf{Copyright infringement determination via multi-LVLM comparison.} 
Many studies \cite{du2023improving, chan2023chateval, lakara2024mad, liu2024groupdebate} have shown that multi-agent debate can effectively improve the reliability of responses generated by large models. At this stage, we utilize $N$ LVLMs $\pi_i (i = 1, 2, …, N)$ to communicate with each other and evaluate overall similarity. In addition, to align with human judgment preferences, we employ few-shot in-context learning \cite{dong2022survey, agarwal2024many} by presenting multiple pairs of images scored by humans as references. Specifically, for a single agent $\pi_i$, given inputs including $x$, $x_{cr}$, the filtered copyright expressions $z^c$, $z^c_{cr}$,  an instruction $p_i$, and the set of human reference images $D_h$ and their corresponding score set $S_h$, the agent is required to output a score $s_i \in [0,1]$, confidence $c_i \in [0,1]$, and supporting rationale $r_i$. Specifically, the process can be represented as:
\begin{equation}
\pi_i(x, x_{{cr}}, z^c, z^c_{cr}, p_i, D_h, S_h) \to (s_i, c_i, r_i).
\end{equation}
Following \cite{du2023improving}, we adopt a fully connected synchronous communication debate approach, where each LVLM receives the responses ($s$, $c$, $r$) from the other $N-1$ LVLMs before making the next judgment. This creates a dynamic feedback loop that strengthens the reliability and depth of the analysis, as models adapt their evaluations based on new insights presented by their peers. Each LVLM can adjust its score based on the responses from the other LVLMs or keep it unchanged. We use the following consistency judgment criterion: 
\begin{equation}
\left| s_i - s_j \right| \leq \alpha \quad \forall i, j \in \{1, 2, \dots, N\}.
\end{equation}
If the difference in scores between all LVLMs is less than $\alpha$, we consider that all models have reached a consensus. Additionally, to avoid the models getting stuck in a meaningless loop, we set the maximum number of debate rounds to $M$.

\begin{figure*}[!ht]
    \centering
    \includegraphics[width=\linewidth]{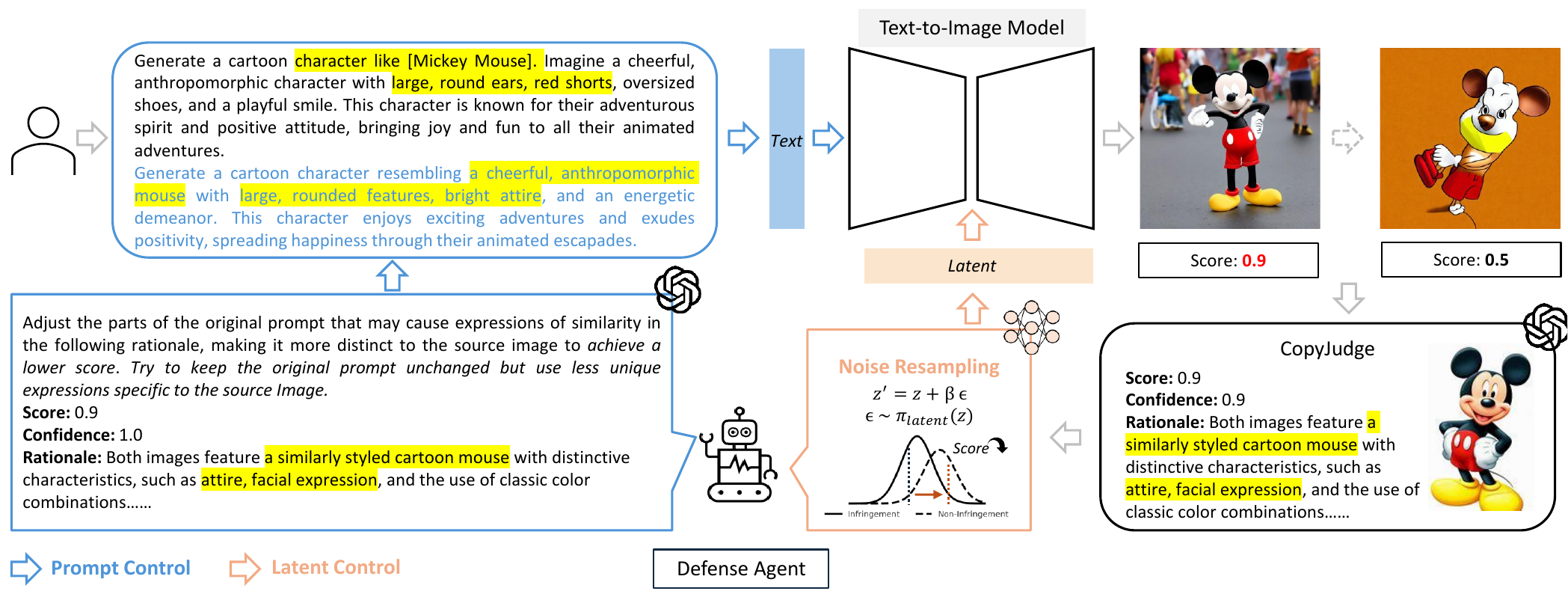}
    \vspace{-3mm}
    \caption{Copyright infringement mitigation framework with controlling input prompts and latent noise. For prompt control, the defense agent iteratively optimizes prompts to mitigate copyright risks in generated images given the CopyJudge's feedback. For latent control, it integrates a RL-based defense agent with reward-guided latent sampling to reduce the predicted infringement scores.}
    \label{fig:mit}
\end{figure*}

After the debate, the agreed-upon results will be input into an independent meta-judge LVLM $\pi_{f}$, which synthesizes the results to give the final score on whether substantial similarity has occurred, defined as: 
\begin{equation}
\pi_{f}(x, x_{{cr}}, z^c, z^c_{cr}, p_{f}, D_h, S_h, S_m, C_m, R_m) \to (s_{f}, c_{f}, r_{f}),
\end{equation}
where $S_m$, $C_m$, and $R_m$ represent the set of scores, confidence levels, and rationales from $N$ LVLMs after reaching consensus in the $m$-th ($m \leq M$) debate. By combining the strengths of individual agents and iterative debating, the approach could achieve a reliable assessment of visual similarity.

Furthermore, we can determine whether the generated image constitutes infringement based on whether the final similarity score exceeds a specific threshold $\gamma$:
\begin{equation}
\text{IsInfringement}(x) = \mathbb{I}(s_{f}) > \gamma.
\end{equation}
The whole two-stage process ensures a comprehensive and reliable evaluation by integrating multiple perspectives and rigorous analysis. Algorithm \ref{alg:infringement_identification} shows the complete process. All instruction prompts can be found in the Appendix \ref{promptA}.

\section{LVLM for Infringement Mitigation}
\label{mitigation}
Based on the identification results, we attempt to control the generation model to ensure its outputs do not infringe on copyright. As shown in Figure \ref{fig:mit}, we will discuss two methods separately depending on the control target: prompt control and latent control.

\subsection{Mitigation via LVLM-based Prompt Control}
\citeauthor{wen2024detecting} have proved that slightly modifying the overfitted prompts can effectively avoid generating memorized images. To achieve automated prompt modification aimed at eliminating infringement expressions, we use an LVLM as a prompt optimizer to iteratively adjust the infringing prompt until the final score falls below a threshold $\gamma$. Formally, given the source image $x_{{cr}}$, the generated image $x^t$ and the corresponding prompt $p^t$ at round $t$, control condition $p^c$, historical judgment score $s^t_{f}$, confidence $c^t_{f}$, and rationale $r^t_{f}$, the prompt modifier $\pi_{p}$ is tasked with providing the prompt for the next round, that is: 
\begin{equation}
\pi_{p}(x^t, x_{{cr}}, p^t, p^c, s^t_{f}, c^t_{f}, r^t_{f}) \to p^{t+1}.
\end{equation}
Here, $p^c$ requires the modifier to alter the infringing expression while preserving the original expression as much as possible to avoid generating meaningless images. This mitigation strategy does not require any knowledge of text-to-image models, making it suitable for general black-box scenarios. The algorithm and instruction prompts can be seen in Algorithm \ref{alg:pc} and Appendix \ref{promptB}.

\begin{algorithm}
\caption{Copyright Infringement Mitigation via Prompt Control}
\label{alg:pc}
\begin{algorithmic}[1]
\REQUIRE Source image $x_{\text{cr}}$, generated image $x^0$, initial prompt $p^0$, control condition $p^c$, LVLM-based prompt modifier $\pi_p$, infringement identification function $f$, threshold $\gamma$, maximum iterations $T$
\STATE Initialize prompt: $p^t \gets p^0$, generated image: $x^t \gets x^0$, iteration counter: $t \gets 0$
\WHILE{$t < T$}
    \STATE Conduct infringement judgement: $s^t, c^t, r^t \gets f(x^t, x_{\text{cr}})$
    \IF{$s^t \leq \gamma$}
        \STATE Break loop
    \ENDIF
    \STATE Generate new prompt: 
    \quad $p^{t+1} \gets \pi_p(x^t, x_{\text{cr}}, p^t, p^c, s^t, c^t, r^t)$
    \STATE Generate new image using $p^{t+1}$: $x^{t+1} \gets \text{T2I}(p^{t+1})$
    \STATE Update iteration counter: $t \gets t + 1$
\ENDWHILE
\STATE \textbf{Return} final prompt $p^t$ and generated image $x^t$
\end{algorithmic}
\end{algorithm}

\begin{algorithm}
\caption{Copyright Infringement Mitigation via Latent Control}
\label{alg:lc}
\begin{algorithmic}[1]
\REQUIRE Initial latent variable $z^0$, source image $x_{\text{cr}}$, generated image $x^0$, initial prompt $p^0$, NN-based policy net $\pi_{\omega}$, infringement identification function $f$, threshold $\gamma$, learning rate $\beta$, maximum iterations $T$
\STATE Initialize latent variable: $z^t \gets z^0$, generated image: $x^t \gets x^0$, iteration counter: $t \gets 0$; Initialize policy parameters $\omega$
\WHILE{$t < T$}
    \STATE Sample latent noise: $\epsilon^t \sim \pi_{\omega}(z^t)$
    \STATE Generate image: $x^t \gets \text{T2I}(z^t, \epsilon^t, p^0)$
    \STATE Conduct infringement judgement: $s^t, c^t, r^t \gets f(x^t, x_{\text{cr}})$
    \IF{$s^t \leq \gamma$}
        \STATE Break loop
    \ENDIF
    \STATE Compute reward: $R(z^t) \gets -\log(s^t)$
    \STATE Compute gradient of the objective and update policy net parameters: 
    \quad $\nabla_{\omega}L(\omega) \gets \mathbb{E}_{z^t \sim \pi_{\omega}}[\nabla_{\omega}\log(\pi_{\omega}) R(z^t)]$
    \STATE Update latent variable: $z^{t+1} \gets z^t + \beta \epsilon^t$
    \STATE Update iteration counter: $t \gets t + 1$
\ENDWHILE
\STATE \textbf{Return} final latent variable $z^t$ and generated image $x^t$
\end{algorithmic}
\end{algorithm}

\subsection{Mitigation via RL-based Latent Control}
For diffusion models, the output is influenced not only by the prompt but also by the latent noise. Latent noise represents encoded representations of the input that capture essential features in a lower-dimensional space. These latent variables guide the generation process, affecting the finer details of the resulting image. In this section, we propose a reinforcement learning (RL)-based latent control method to mitigate copyright infringement in diffusion-based generative models. Our method involves training an agent to search the input latent variables that yield lower infringement scores, ensuring that the generated outputs do not violate copyright.

Specifically, for latent variable \( z \), we define a policy \( \pi_{\omega} \) parameterized by \( \omega \), allowing us to sample latent noise \( \epsilon \sim \pi_{\omega}(z) \), which follows a Gaussian distribution. The sampled noise \( \epsilon \) is then passed through the pre-trained diffusion decoder \( f \) to produce the image \( x = f(z, \epsilon) \).

\begin{figure*}[!ht]
    \centering
    \captionsetup[subfigure]{skip=0.1pt}
    \subfigure[Distribution on D-Rep Dataset]{ 
        \includegraphics[width=\linewidth]{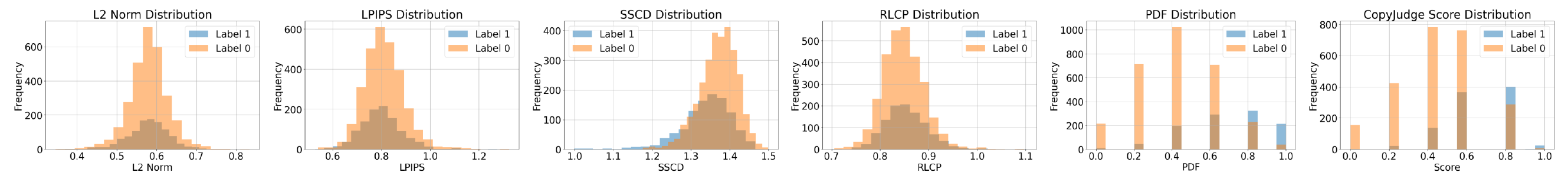}
        \label{fig:res1_drep}
    }
    \subfigure[Distribution on Cartoon Character Dataset]{
        \includegraphics[width=\linewidth]{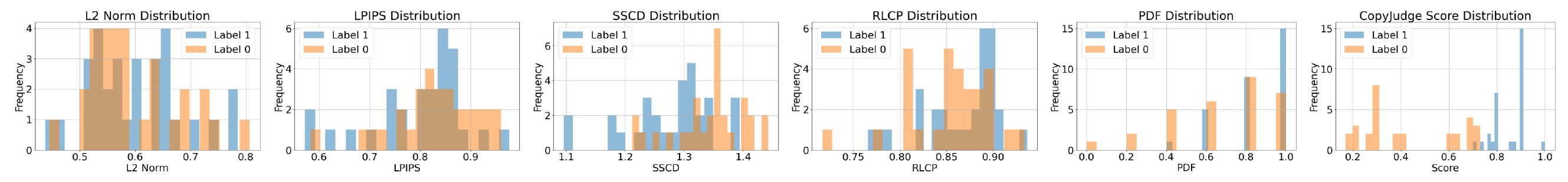}
        \label{fig:res1_cartoon}
    }
    \subfigure[Distribution on Artwork Dataset]{
        \includegraphics[width=\linewidth]{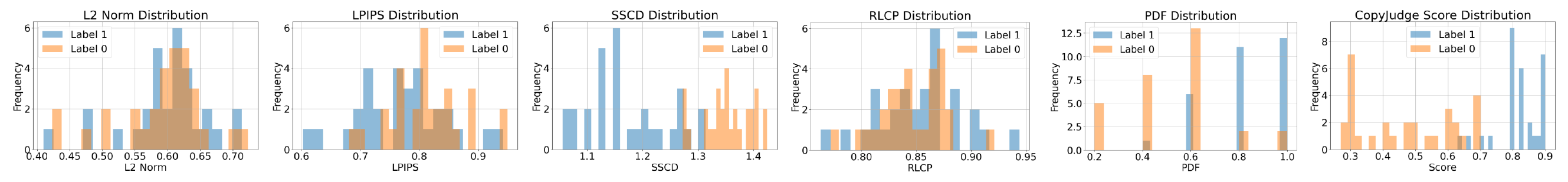}
        \label{fig:res1_artwork}
    }
    \caption{Prediction score distributions for infringing and non-infringing samples across different infringement identification metrics.}
    \label{fig:res1}
\end{figure*}

To assess the copyright infringement potential of the generated image, we employ our \textit{CopyJudge} to obtain the infringement score $s_{f}$. Based on this score, we define a reward function:
\begin{equation}
    R(z) = -\log(s_{f}).
\end{equation}
This reward is designed to penalize outputs with higher infringement scores, thus encouraging the generation of non-infringing content. We optimize the parameters \( \omega \) by maximizing the expected reward, \( L(\omega) \), defined as:
\begin{equation}
    L(\omega) = \mathbb{E}_{z \sim \pi_{\omega}}[R(z)].
\end{equation}
The gradient of this objective is computed using the REINFORCE rule \cite{williams1992simple}, which is given by:
\begin{equation}
\nabla_{\omega}L(\omega) = \mathbb{E}_{z \sim \pi_{\omega}}[\nabla_{\omega}\log(\pi_{\omega}) R(z)].
\end{equation}
During the training process, the latent variable $z$ is updated according to the following rule:
\begin{equation}
 z' = z + \beta \epsilon, \quad \epsilon \sim \pi_{\omega}(z),
\end{equation}
where $\beta$ is the step size. We further conduct normalization for the latent variables to maintain stability and prevent extreme deviations. This  RL-based approach allows the agent to explore variations in the latent space, thereby improving its ability to generate non-infringing content. Algorithm \ref{alg:lc} shows the detailed process.

\section{Experiments}
\label{experiments}
\subsection{Infringement Identification Experiments}
\label{iden setup}
\textit{Dataset.} Firstly, we use D-Rep dataset \cite{wang2024image}, which contains realistic and generated image pairs scored by human from 0 to 5 according to their similarity. Following its setting, we consider samples with scores of 4 or above as infringement samples, while the rest were non-infringement samples. We use the 4,000 official test images. In addition, we also consider the specific IP infringement. Referring to \cite{ma2024dataset}, we select 10 well-known cartoon characters and 10 artworks from Wikipedia (all can be found in the Appendix \ref{promptB}). Three different text-to-image models—Stable Diffusion v2, Kandinsky2-2 \cite{razzhigaev2023kandinsky}, and Stable Diffusion XL \cite{podell2023sdxl}—are used to generate images. For each item, we manually select one infringing image and one non-infringing image from each model, resulting in a total of 60 positive samples and 60 negative samples. 

\textit{Baselines.} We select 4 commonly used distance-based image copy detection metrics: $L_2$ norm \cite{carlini2023extracting}, which directly measures pixel-wise differences; LPIPS \cite{zhang2018unreasonable}, which captures perceptual similarity based on deep network features; SSCD \cite{pizzi2022self}, which learns transformation-invariant representations to match generated images with their original training counterparts; and RLCP \cite{shi2024rlcp}, which integrates semantic and perceptual scores for a more comprehensive similarity assessment. Additionally, we compare our approach with the state-of-the-art image copy detection method PDF-Emb \cite{wang2024image}, which models the similariy level of image pairs as a probability density function. We implement all methods using their official open-source code.

\textit{Metric.} Accuracy and F1 score are calculated as the criteria for infringement classification. We perform grid search to select the threshold that achieves the highest F1 score for each method.

\textit{Implementation detail.} Although our approach does not rely on a specific LVLM, we choose GPT-4o \cite{hurst2024gpt} as our agent by default, which has been proven in \cite{xu2025can} to achieve the best infringement detection performance compared to other LVLMs. As for multi-agent debate, we use 3 agents with a maximum of 5 iterations. We use random 3 images from each level (0-5) in the D-Rep training set as human priors to present to the agent.

\begin{table}[t]
\caption{Infringement identification results on various dataset (Accuracy $\uparrow$ / F1-Score$\uparrow$ $\times 100$).}
\vspace{-0.2in}
\label{table1}
\vskip 0.15in
\begin{center}
\begin{tabular}{l@{\hskip 4pt}c@{\hskip 4pt}c@{\hskip 4pt}c}
\toprule
{Method / Dataset} & D-Rep  & Cartoon & Artwork\\
\midrule
$L_2$ norm & 27.42 / 42.32 &51.67 / 67.42 & 50.00 / 66.67\\
LPIPS &26.85 / 42.29&56.67 / 68.29 &66.67 / 67.74\\
SSCD    &43.65 / 44.37&58.33 / 70.59 & 90.00 / 90.32\\
RLCP    &27.02 / 42.23& 50.00 / 66.67 &50.00 / 66.67\\
PDF-Emb    &\textbf{79.90} / \textbf{57.10}& 61.67 / 71.60 & 81.67 / 80.70\\
{CopyJudge (Ours)}  &75.67 / 51.04 & \textbf{91.67} / \textbf{90.91} & \textbf{93.33} / \textbf{92.86}\\
\bottomrule
\end{tabular}
\end{center}
\end{table}

\textbf{Results.} From the results in Table \ref{table1}, it is evident that traditional image copy detection methods exhibit limitations in the copyright infringement identification task. Our approach significantly outperforms most methods. For the state-of-the-art method, PDF-Emb, which was trained on 36,000 samples from the D-Rep, our performance on D-Rep is slightly inferior. However, its poor performance on the Cartoon IP and Artwork dataset highlights its lack of generalization capability, whereas our method demonstrates equally excellent results across datasets. Figure \ref{fig:res1} illustrates the prediction score distributions for all methods, showing that our approach achieves a relatively more distinct boundary between infringing and non-infringing cases. More results and ablation studies can be found in the Appendix \ref{resultsA} and \ref{ablation}.


\subsection{Infringement Mitigation Experiments}
To thoroughly test the effectiveness of our infringement mitigation, we consider both memorization and specific IP infringement mitigation. 

\subsubsection{Memorization Mitigation}
\citeauthor{wen2024detecting} have found that text-to-image model memorization can be mitigated through simple prompt modifications without additional model training. They iteratively adjust prompt embeddings through gradients based on text-conditional noise prediction to reduce memorization. To compare with it, we conduct our memorization mitigation test using the overfitted Stable Diffusion v1 model trained by them. We use the same 200 memorized images provided by them and record the average Infringement Score identified by \textit{CopyJudge} before and after the mitigation. At the same time, we use the CLIP Score \cite{radford2021learning} to measure the alignment between the modified images and the original prompt. Since \citeauthor{wen2024detecting} have demonstrated that the generated copy images do not change with variations in latent space, we only use the prompt control (PC) method here. Both \citeauthor{wen2024detecting}’s and our method are set to perform 10 prompt optimization iterations.

\begin{figure}[!ht]
    \centering
    \includegraphics[width=\linewidth]{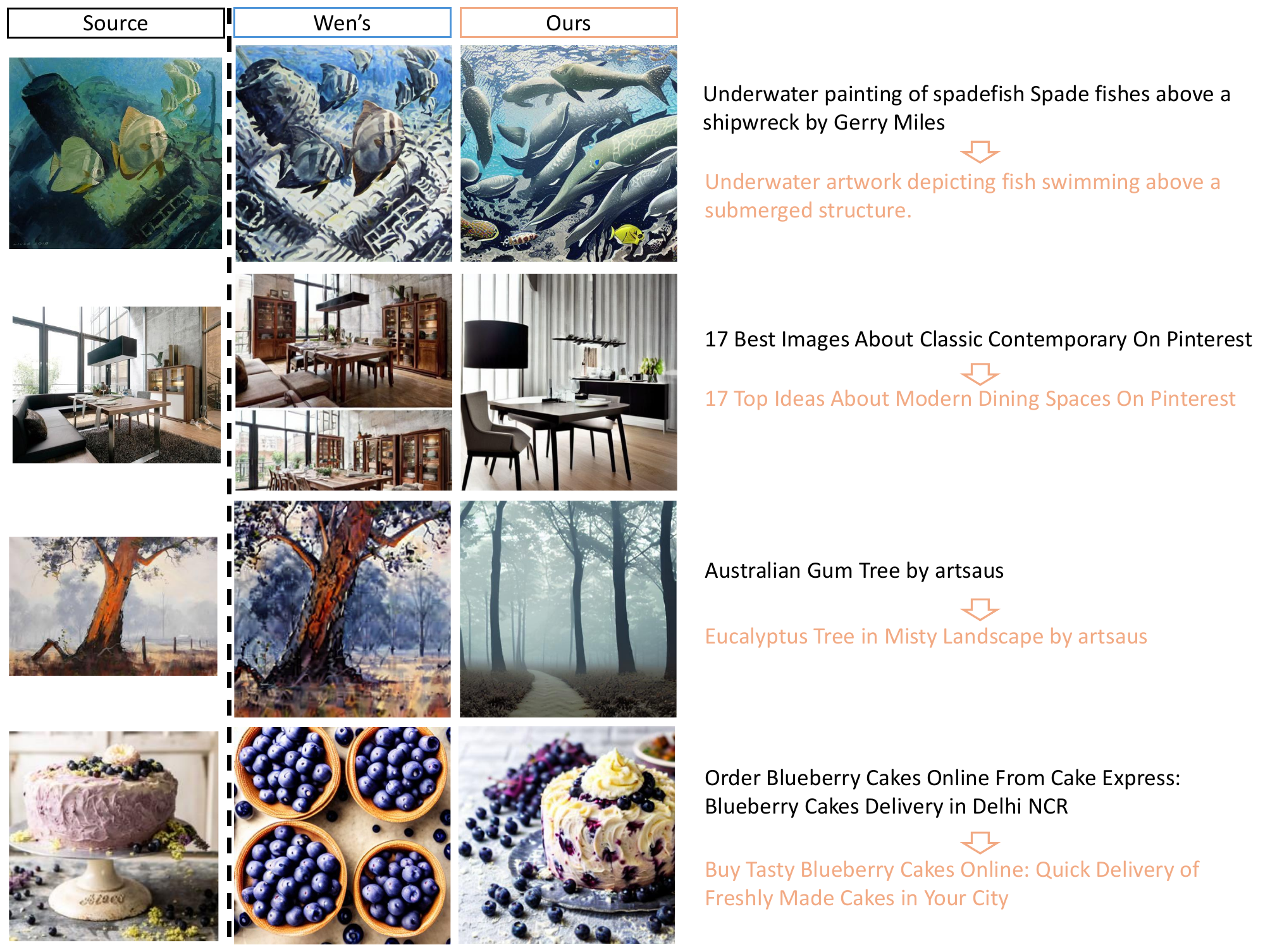}
    \vspace{-3mm}
    \caption{Visualization of generated images and corresponding prompts before and after memorization mitigation.}
    \label{fig:res2}
    \vspace{-4mm}
\end{figure}

\textbf{Results.}
From Table \ref{table3}, our approach could generate images that are less likely to cause infringement while maintaining a comparable, slightly reduced match accuracy. As shown in Figure \ref{fig:res2}, our method effectively avoids the shortcomings of \citeauthor{wen2024detecting}’s method, including failing to mitigate memorization or generating highly deviated images.

\begin{table}[t]
\caption{Memorization mitigation results (Infringement Score $\downarrow$ / CLIP Score $\uparrow$).}
\vspace{-0.1in}
\label{table3}
\vskip 0.15in
\begin{center}
\begin{tabular}{lcc}
\toprule
{Method} & Infringement Score & CLIP Score \\
\midrule
Raw    &0.869&31.65\\
\citeauthor{wen2024detecting}'s Method  &0.605&28.96\\
PC (ours) &\textbf{0.468}&28.31\\
\bottomrule
\end{tabular}
\end{center}
\vskip -0.15in
\end{table}

\subsubsection{IP Infringement Mitigation}
Compared to cases of exact replication (memorization), here we consider specific IP infringement, such as imitation of cartoon IPs and artistic elements. Based on whether the input prompt contains direct copyright information, we consider two types of infringement scenarios: \textit{explicit infringement} and \textit{implicit infringement}.

\textbf{Explicit infringement}. This refers to prompts that directly contain copyright information, such as \textit{“Generate an image of Mickey Mouse.”} We use the 20 cartoon and artwork samples collected in section \ref{iden setup} to generate infringing images using Stable Diffusion v2, where the prompt explicitly includes the names or author names of the work.

\textbf{Implicit infringement.} This occurs when the prompt does not explicitly contain copyright information, but the generated image still infringes due to certain infringing expressions. This type of scenario is more applicable to commercial text-to-image models, as they often include content detection modules that can effectively detect copyrighted information and thus reject the request. In this scenario, we use the same IP samples as above, but generate infringing images without any explicit copyright information using DALL·E 3 \cite{Betker2023ImprovingImageGeneration}, which has a safety detection module to reject prompts that trigger it.

\textbf{Automated attack.} Efficiently retrieving or generating infringing prompts has always been a challenge. \citeauthor{kim2024automatic} utilize large models to iteratively generate jailbreak prompts targeting commercial models, thereby inducing them to output copyrighted content. Drawing from it, we use our \textit{CopyJudge} to generate infringing prompts. In contrast to mitigation, for attack, we only need to use an LVLM to progressively intensify the infringing expressions within the prompt. The prompt is iteratively adjusted, and once the infringement score exceeds 0.8 / 1.0, mitigation is activated, using the current prompt as the starting point.

\begin{figure*}[!ht]
    \centering
    \includegraphics[width=\linewidth]{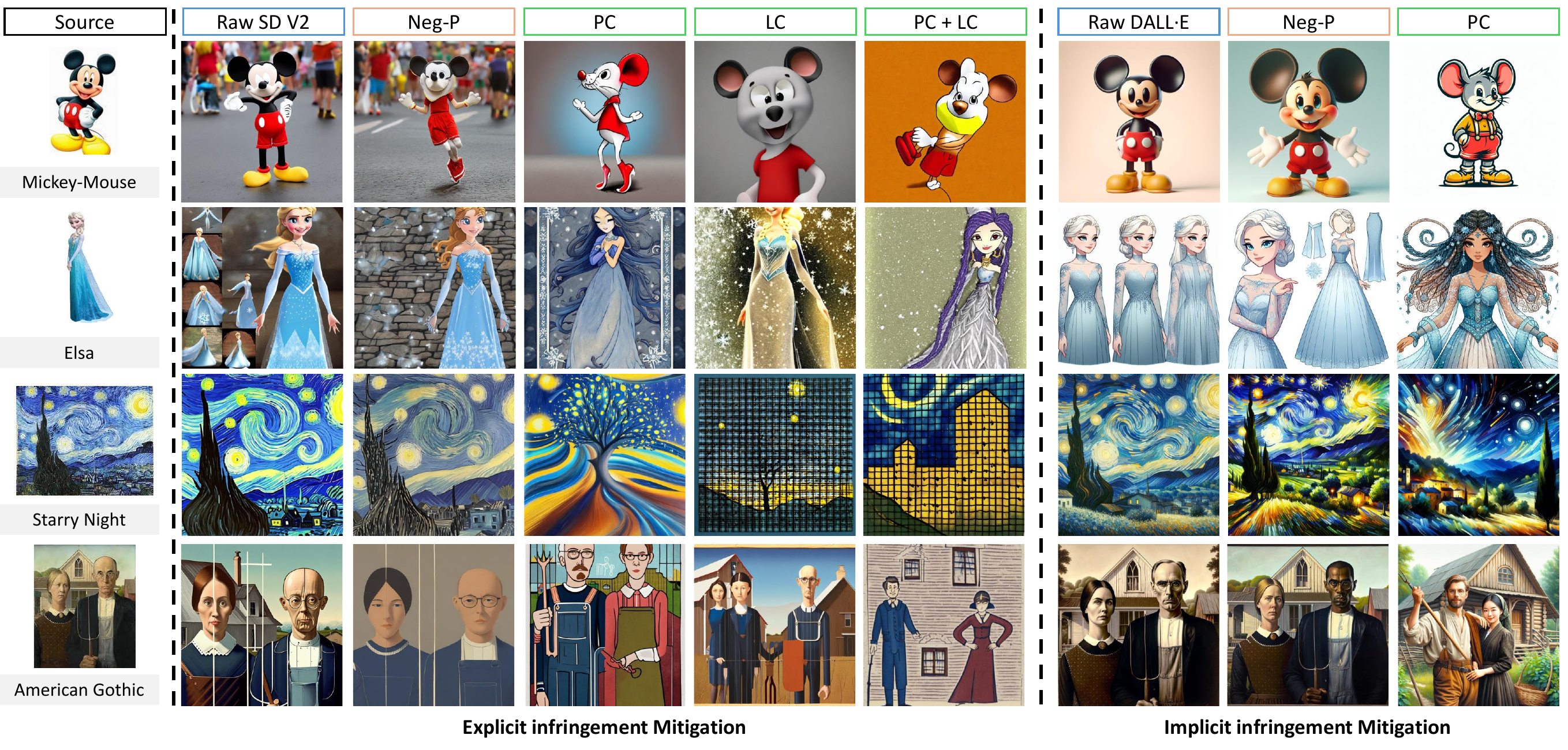}
    \vspace{-7mm}
    \caption{Visualization of generated images before and after IP infringement mitigation.}
    \label{fig:res3}
    \vspace{-0.1in}
\end{figure*}

\begin{table}[t]
\caption{Explicit IP infringement mitigation results on Stable Diffusion v2 (Infringement Score $\downarrow$ / CLIP Score $\uparrow$).}
\vspace{-0.2in}
\label{table4}
\vskip 0.15in
\begin{center}
\begin{tabular}{l@{\hskip 30pt}c@{\hskip 30pt}c}
\toprule
{Method / Dataset} & Cartoon & Artwork\\
\midrule
Raw    &0.810 / 32.39 &0.760 / 35.21 \\
Neg-P  &0.597 / 30.91  & 0.616 / 32.07\\
LC (Ours) &0.673 / 33.19  &0.665 / 33.50\\
PC (Ours) &0.450 / 29.48 &0.417 / 29.46 \\
PC + LC (Ours) &\textbf{0.350} / 28.67 & \textbf{0.353} / 28.82 \\
\bottomrule
\end{tabular}
\end{center}
\end{table}

\begin{table}[t]
\caption{Implicit IP infringement mitigation results on DALL·E 3 (Infringement Score $\downarrow$ / CLIP Score $\uparrow$).}
\vspace{-0.2in}
\label{table5}
\vskip 0.15in
\begin{center}
\begin{tabular}{l@{\hskip 30pt}c@{\hskip 30pt}c}
\toprule
{Method / Dataset} & Cartoon & Artwork\\
\midrule
Raw    &0.783 / 30.54 & 0.697 / 29.80\\
Neg-P &0.710 / 29.27 & 0.689 / 29.34\\
PC (Ours) &\textbf{0.547} / 27.92&\textbf{0.431} / 28.81\\
\bottomrule
\end{tabular}
\end{center}
\vskip -0.1in
\end{table}

For explicit infringement, we validate both prompt control (PC) and latent control (LC). For implicit infringement, due to the commercial model DALL·E's inability to customize latents, we only evaluate prompt control. In both scenarios, we compare our approach with the only prompt-based IP infringement mitigation method \cite{wang2024evaluating}, which detects potentially infringed works using LVLM and inputs the detected work information as a negative prompt (Neg-P) into the generative model to avoid infringement. We follow the same settings as in the original paper, but the paper does not address how to handle commercial models that cannot accept negative prompts. For such models, we simply appended the suffix "without [copyrighted work information (e.g., name, author, etc.)]" to the original prompt to simulate a negative prompt.

\textbf{Results.}
From Tables \ref{table4} and \ref{table5}, it can be seen that our method significantly reduces the likelihood of infringement, both for explicit and implicit infringement, with only a slight drop in CLIP Score. The infringement score after only latent control is relatively higher than after prompt control because retrieving non-infringing latents without changing the prompt is quite challenging. However, we can still effectively reduce the infringement score while maintaining higher image-text matching quality. Figure \ref{fig:res3} shows visualization results, where it can be observed that we avoid the IP infringement while preserving user requirements. Additional results are provided in the Appendix \ref{resultB}.


\section{Conclusion and Future Work}
\label{conclusion}
We present \textit{CopyJudge}, an innovative framework for automating the identification of copyright infringement in text-to-image diffusion models. By leveraging abstraction-filtration-comparison test and multi-LVLM debates, our approach could effectively evaluate the substantial similarity between generated and copyrighted images, providing clear and interpretable judgments. Additionally, our LVLM-based mitigation strategy helps avoid infringement by automatically optimizing prompts and exploring non-infringing latent noise vectors, while ensuring that generated images align with the user's requirements.

While our method demonstrates encouraging results, its current performance is influenced by the limited availability of high-quality labeled data. This presents an opportunity for further refinement, particularly in developing a more comprehensive dataset that reflects diverse and nuanced human judgment criteria. We also anticipate future work involving stronger and more varied adversarial cases, which would allow us to better assess the robustness of our identification and mitigation strategies under more challenging conditions. Moreover, although our current framework is model-agnostic and has primarily utilized GPT-based LVLMs, we are interested in exploring a broader range of models in the future to enhance the generalizability and effectiveness of our approach in safeguarding copyright.

\section*{Acknowledgements}
We gratefully acknowledge the financial and IT support provided by the EPFL Artificial Intelligence Laboratory.

\clearpage
\bibliographystyle{ACM-Reference-Format}
\bibliography{acmart}

\clearpage
\appendix
\section{Supplementary Information on Infringement Identification}

\subsection{Instructive Prompts Used}
\label{promptA}
The effectiveness of the infringement identification process relies on well-designed prompts. Below are examples of instructive prompts used in our framework:

\begin{itemize}
    \item \textbf{Abstraction Prompt:}  
    ``Please help decompose the given two images into their abstract concepts. Identify key components such as composition, themes, color palette, textures, and any unique visual elements. Describe each component in detail.  
    Ensure the output follows the template format: `Image1: XXX, Image2: XXX'.''

    \item \textbf{Filtration Prompt:}  
    ``Based on the image decomposition, please identify which elements are likely unprotectable under copyright law. These could include generic concepts, common patterns, functional aspects, or elements derived from public domain sources. Then, specify the remaining unique elements that are original and creative.  
    Ensure the output follows the template format: `Image1 Unique Elements: XXX, Image2 Unique Elements: XXX'.''

    \item \textbf{Comparison and Scoring Prompt:}  
    ``Please assign an overall similarity score (0–1) between the two images based on the unique elements. The score should be higher if there are similar elements. Provide a confidence score on a scale of 0 to 1, where 1 indicates very high confidence in the evaluation. Additionally, provide a rationale for whether this case constitutes likely copyright infringement. Ensure the output follows the template format: `Score: [0-1], Confidence: [0-1], Reason: [clear and concise explanation]'.''

    \item \textbf{Agent Feedback Integration Prompt:}  
    ``The following is feedback from other agents: 
    
    \textit{Agent 1 score: S1, confidence: C1, reason: R1}  
    
    \textit{Agent 2 score: S2, confidence: C2, reason: R2}

    ……(if any)
    
    You may adjust your score based on this information or maintain your judgment. Ensure the output follows the template format:  `Score: [0-1], Confidence: [0-1], Reason: [clear and concise explanation]'.''

    \item \textbf{Final Decision Prompt:}  
    ``The original instruction is: [comparison and scoring prompt]. Below are the scores, confidence levels, and rationales from multiple judges. Please combine them to make a final decision.  
    
    Scores: [S1, S2, S3, ……(if any)]
    
    Confidences: [C1, C2, C3, ……(if any)] 
    
    Rationales: [R1, R2, R3, ……(if any)]  
    
    Ensure the output follows the template format: `Score: [0-1], Confidence: [0-1], Reason: [clear and concise explanation]'.''

\end{itemize}

\subsection{Ablation Studies}
\label{ablation}
We separately explore the contributions of the abstraction-filtration-comparison test (AFC), multi-agent debate (MAD), and prior demonstration (DEM) to infringement identification. We randomly select 200 samples from the D-Rep dataset for ablation experiments, with each group undergoing five independent runs. As shown in Table \ref{table6}, each module has a positive impact. Specifically, abstraction-filtration-comparison significantly improves accuracy, the demonstration of human priors effectively enhances the F1 score, and the multi-agent debate further boosts both metrics, ensuring reliable identification results.

\begin{table}[t]
\caption{Ablation studies on different identification modules.}
\vspace{-0.2in}
\label{table6}
\vskip 0.15in
\begin{center}
\begin{tabular}{c@{\hskip 10pt}c@{\hskip 10pt}c@{\hskip 10pt}c|c}
\toprule
LVLM & AFC  & MAD & DEM & ACC / F1-Score $\uparrow$  \\
\cmidrule(lr){1-5}
\checkmark & - & - & - & 73.70 $\pm$ 0.68 / 42.44 $\pm$ 1.08  \\
\checkmark & \checkmark & - & - & 76.50 $\pm$ 1.70 / 44.79 $\pm$ 2.30\\
\checkmark & - & \checkmark & - & 75.00 $\pm$ 1.14 / 43.48 $\pm$ 1.57 \\
\checkmark & - & - & \checkmark & 76.50 $\pm$ 1.14 / 48.35 $\pm$ 1.53 \\
\checkmark & \checkmark & \checkmark & - & 77.50 $\pm$ 1.70 / 47.62 $\pm$ 1.81 \\
\checkmark & \checkmark & \checkmark & \checkmark & \textbf{78.50} $\pm$ 2.00 / \textbf{51.70} $\pm$ 2.19\\
\bottomrule
\end{tabular}
\end{center}
\end{table}

\subsection{Additional Results}
\label{resultsA}
\begin{figure*}[!ht]
    \centering
    \includegraphics[width=\linewidth]{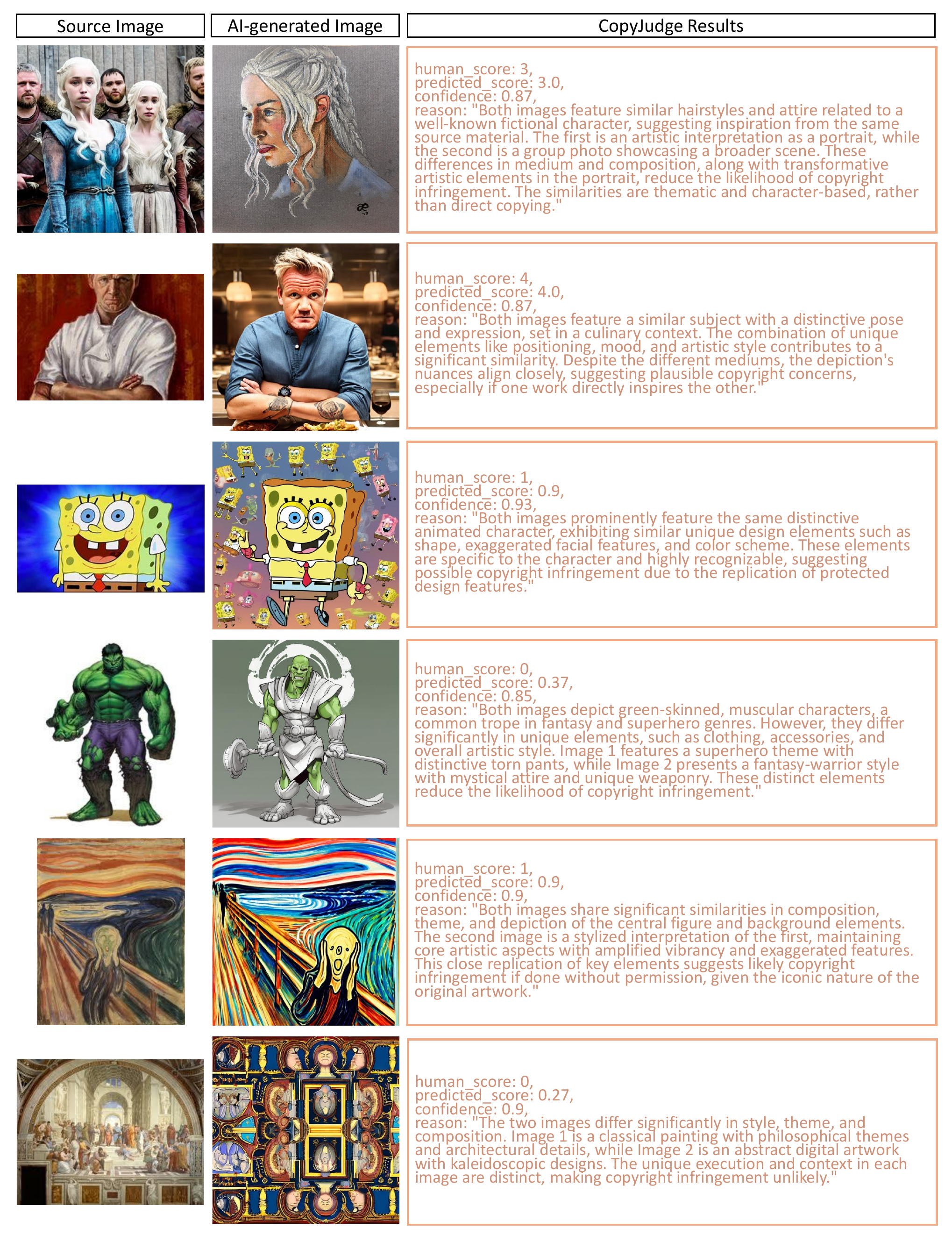}
    \vspace{-6mm}
    \caption{Successful copyright infringement identification by CopyJudge.}
    \label{fig:app1}
    \vspace{-3mm}
\end{figure*}

\begin{figure*}[!ht]
    \centering
    \includegraphics[width=\linewidth]{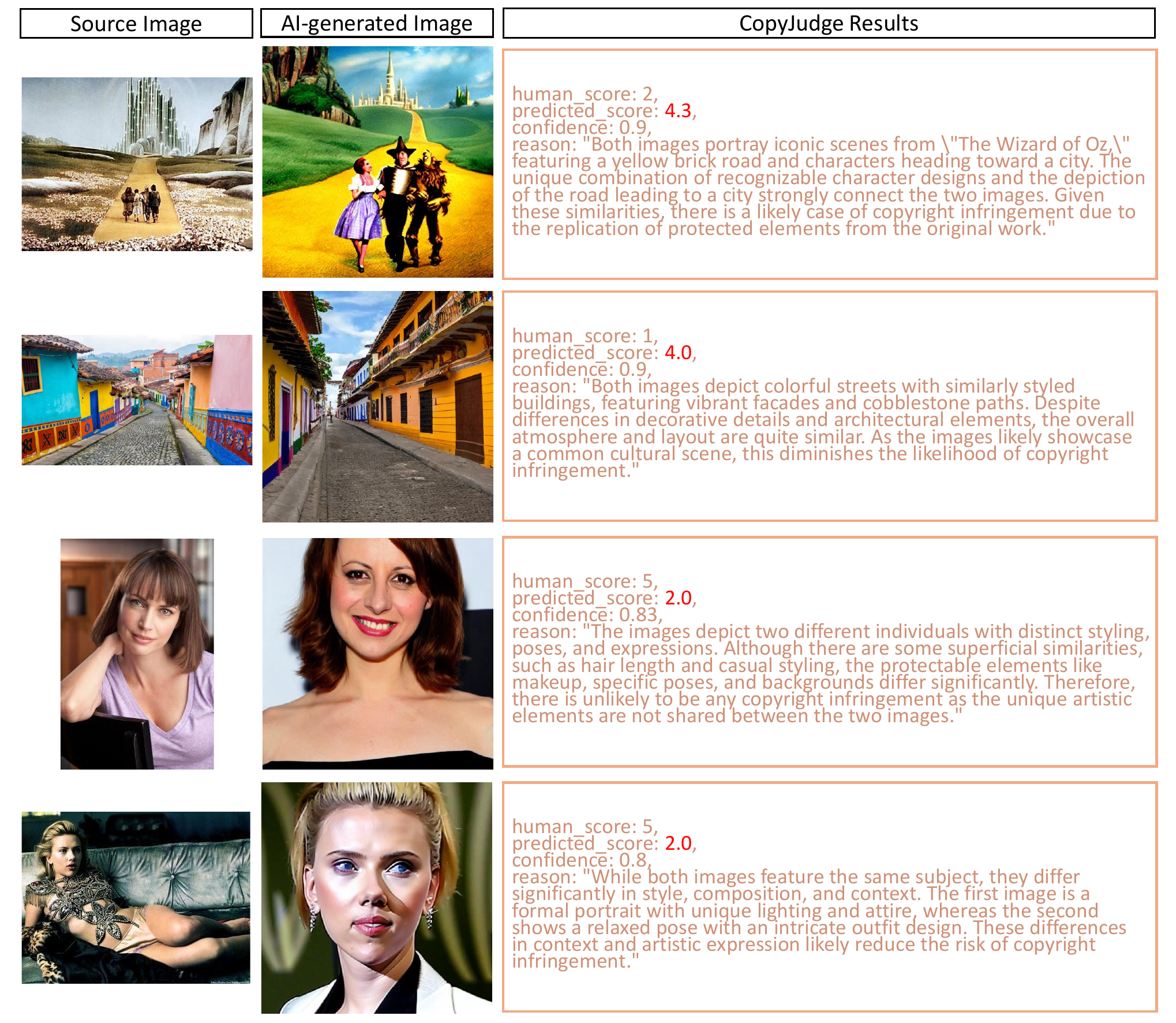}
    \vspace{-6mm}
    \caption{Failure case of copyright infringement identification by CopyJudge.}
    \label{fig:app2}
    \vspace{-3mm}
\end{figure*}

We present some successful cases of copyright infringement identified by our framework in Figure \ref{fig:app1}, while Figure \ref{fig:app2} shows some failure cases. We found that these failures mainly stem from excessive sensitivity to specific compositions and imprecise recognition of real individuals.


\textit{Time Consumption.} The computational cost of running the \textit{CopyJudge} is an important consideration. Table~\ref{tab:time} provides the average runtime per sample for different steps in the process. We call the official API for testing and find that a single API call, along with extracting useful information, takes approximately 12s. The main time overhead comes from the multi-agent debate process. We acknowledge that this could be a limitation of our approach, while it allows for higher accuracy. In the future, we aim to improve the deployment and interaction of LVLMs to achieve faster decision-making.

\begin{table*}[h]
    \centering
    \begin{tabular}{c|c}
        \hline
        Process Step & Average Time (seconds) \\
        \hline
        Single GPT-4o  & 12.3 \\
        Copyright Expression Extraction (Abstraction + Filtration)  & 13.1 \\
        Copyright Infringement Determination (Multi-LVLM Comparison) & 75.6 \\
        Whole Process & 88.7 \\
        \hline
    \end{tabular}
    \caption{Time analysis of the infringement identification process.}
    \label{tab:time}
\end{table*}

\begin{table*}[h]
    \centering
    \begin{tabular}{c|c}
        \hline
        Process Step & Average Time (seconds) \\
        \hline
        Infringement Identification  & 88.7 \\
        Latent Control  & 92.5 \\
        Prompt Control  & 101.7\\
        Latent Control + Prompt Control & 115.5 \\
        \hline
    \end{tabular}
    \caption{Time analysis of the infringement mitigation process.}
    \label{tab:time2}
\end{table*}

\begin{table*}[h]
    \centering
    \begin{tabular}{c|c}
         \hline
    \textbf{Author/Creator} & \textbf{IP Name} \\
    \hline
    Disney & Elsa, Mickey Mouse \\
    Nickelodeon / Paramount Global & SpongeBob SquarePants \\
    Sanrio & Hello Kitty \\
    Marvel / Disney & Iron Man, Hulk, Spider-Man \\
    DC Comics / Warner Bros. & Batman \\
    Nintendo & Super Mario \\
    Akira Toriyama (Toei Animation / Shueisha) & Goku (Dragon Ball) \\
    \hline
    Leonardo da Vinci & Mona Lisa \\
    Vincent van Gogh & Starry Night \\
    Edvard Munch & The Scream \\
    Sandro Botticelli & The Birth of Venus \\
    Johannes Vermeer & The Girl with a Pearl Earring \\
    Gustav Klimt & The Kiss \\
    Eugène Delacroix & Liberty Leading the People \\
    Grant Wood & American Gothic \\
    Edward Hopper & Nighthawks \\
    Raphael & The School of Athens \\
    \hline
    \end{tabular}
    \caption{Summary of IPs by author or creator.}
    \label{tab:ip}
\end{table*}

\section{Supplementary Information on Infringement Mitigation}
\subsection{Instructive Prompts and IP Used}
\label{promptB}
We list the Instructive Prompts used below, and in Table \ref{tab:ip}, we list the cartoon characters and artworks used, along with their respective IP ownership by individuals/organizations.

\begin{itemize}
    \item \textbf{Prompt for Initial Attack:} ``Generating a prompt that describes the [IP type (cartoon character / artwork)] [IP Name] in the given image. Ensure the output follows the template format: `Prompt: Generate a [IP type] like [IP name]. XXX'." (For implicit infringement, we delete the IP name.)

    \item \textbf{Modifying a Prompt to Improve Similarity Score (Attack Iteration):}  
    "Adjust the parts of the original prompt of the second image that may cause expressions of distinction in the following rationale, making it more similar to the first image to achieve a higher score. Add more information about the [IP type] in Image 1, and provide more unique expressions specific to the [IP type] in Image 1. You can make any changes as long as they improve the similarity score. 
    
    Original prompt: [Original Prompt]
    
    Judgment results: score-[Score], confidence-[Confidence], reason-[Reason]
    
    Ensure the output follows the template format: `Modified Prompt: Generate a [IP type] like [IP Name]. xxx'." (For implicit infringement, we delete the IP name.)

    \item \textbf{Modifying a Prompt to Reduce Similarity Score (Defense Iteration):}  
    ``Adjust the parts of the original prompt of the second image that may cause expressions of similarity in the following rationale, making it more distinct to the first image to achieve a lower score.
    
    Original prompt: [Original Prompt]
    
   Judgment results: score-[Score], confidence-[Confidence], reason-[Reason]
   
   Try to keep the original prompt unchanged but use less unique expressions specific to Image 1. Ensure the output follows the template format:`'Modified Prompt: xxx'."

\end{itemize}

\subsection{Additional Results}
\label{resultB}
\begin{figure*}[!ht]
    \centering
    \includegraphics[width=\linewidth]{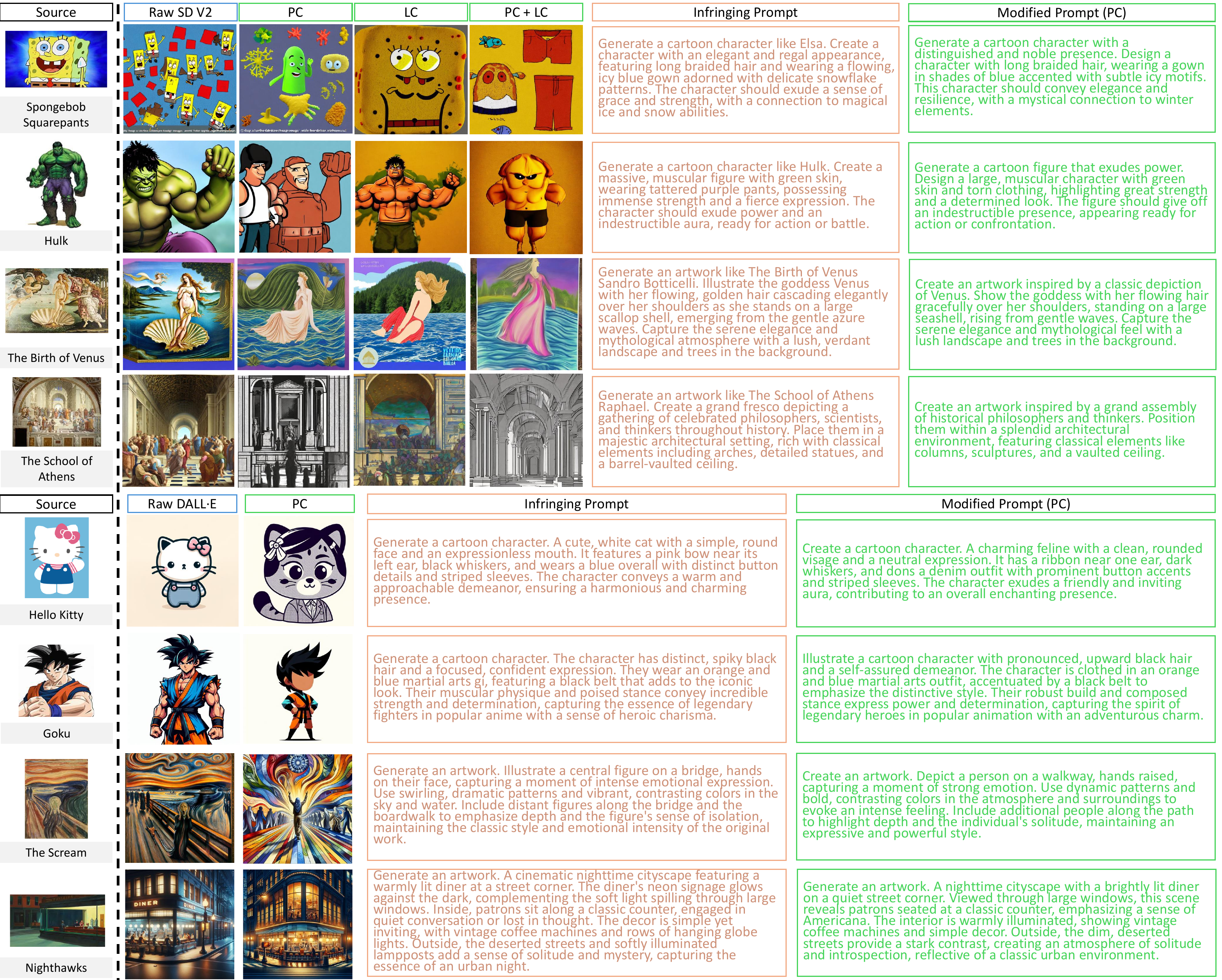}
    \vspace{-6mm}
    \caption{Generated images and corresponding prompts for copyright infringement mitigation.}
    \label{fig:app3}
    \vspace{-3mm}
\end{figure*}
Figure \ref{fig:app3} presents additional results using our prompt control and latent control for mitigating infringement. It can be observed that our approach effectively prevents the generation of infringing images while ensuring closeness to the original expressions.

\textit{Time Consumption.} Table \ref{tab:time2} presents the time consumption for a single mitigation iteration. As shown, the majority of the time is spent on the infringement identification phase (detailed results can be found in Table \ref{tab:time}). In the future, we will focus on improving the efficiency of both identification and mitigation processes.

\end{document}